\documentclass[sigconf]{acmart}

\setcopyright{none}
\renewcommand\footnotetextcopyrightpermission[1]{}
\usepackage[dvipsnames]{xcolor}
\usepackage{graphicx}
\usepackage{subcaption}
\newcommand{\kibitz}[3]{\textcolor{#1}{[\textbf{#2}: #3]}}
\usepackage{algorithm}
\usepackage{algpseudocode}
\newcommand{\shishir}[1]{\kibitz{red}{SD}{#1}}

\newcommand{\hojat}[1]{\kibitz{ForestGreen}{HA}{#1}}
\newcommand{\added}[1]{\textcolor{MidnightBlue}{#1}} % camera-ready additions to body text; set to {#1} for final
\acmConference[KDD2026 Workshop]{KDD 2026 Workshop on Evaluation and Trustworthiness of Agentic AI}{August 2026}{Jeju, Korea}

\usepackage{tikz}
\usetikzlibrary{arrows.meta,positioning,shapes.geometric,fit,backgrounds}
\usepackage{booktabs}
\begin{document}

\title{Autoresearch for Marketplace Catalogs:\\
From Legacy Forms to AI-Native Matching}
\titlenote{Accepted at the KDD Workshop on Evaluation and Trustworthiness of Agentic AI (KDD 2026).}

\author{Kartik Ravisankar}
\email{kravisankar@thumbtack.com}
\affiliation{%
  \institution{Thumbtack}
  \city{San Francisco}
  \state{CA}
  \country{USA}
}

\author{Hojat Abdolanezhad}
\email{habdolanezhad@thumbtack.com}
\affiliation{%
  \institution{Thumbtack}
  \city{San Francisco}
  \state{CA}
  \country{USA}
}

\author{Daniel Capo}
\email{dcapo@thumbtack.com}
\affiliation{%
  \institution{Thumbtack}
  \city{San Francisco}
  \state{CA}
  \country{USA}
}

\author{Sang Su Lee}
\email{psulee@thumbtack.com}
\affiliation{%
  \institution{Thumbtack}
  \city{San Francisco}
  \state{CA}
  \country{USA}
}

\author{Shishir Dash}
\email{shishirdash@thumbtack.com}
\affiliation{%
  \institution{Thumbtack}
  \city{San Francisco}
  \state{CA}
  \country{USA}
}

\author{Vijay Anand Raghavan}
\email{vraghavan@thumbtack.com}
\affiliation{%
  \institution{Thumbtack}
  \city{San Francisco}
  \state{CA}
  \country{USA}
}

\begin{abstract}
Two-sided service marketplaces that match consumers to service providers are transitioning from deterministic request-form intake to an AI-native probabilistic matching system, enabled by recent advances in large language models (LLMs) that can infer user intent, contextual preferences, and latent constraints from natural language instructions. As marketplaces increasingly rely on inferred intent rather than fixed-form fields, these platforms must regenerate the provider-side preference taxonomy that underwrites matching, search, and pricing: a set of provider attributes and preferences interpretable to service providers while remaining useful signals for marketplace decision-making.
%This transition requires regenerating the pro-side preference-tag catalog that underwrites matching, search, and pricing: a coherent set of tags per trade, with each tag legible to a working pro and meaningful as a screening signal. %
We present an autoresearch loop that generates this taxonomy one occupation at a time. The system has been deployed in production at a major U.S. consumer services marketplace since April 2026 across 132 occupations. Instead of constructing a single global hierarchy, the loop treats each occupation as an independent generation problem and runs iterative propose-evaluate-keep refinement cycles. Each candidate tag set is scored using a recalibrated six-rubric LLM-as-judge framework, producing a composite score of up to 15. A 7-critic panel, each with distinct personas, contributes weighted penalties to produce an adjusted score, with no hard vetoes. A separate LLM-based parity-mapping stage maps legacy request-form Q\&A pairs back to the generated taxonomy, producing both a coverage signal and a scalable interface for human quality assurance. The loop migrates legacy structured Q\&A by first inferring the underlying provider attribute that each question was intended to measure, rather than performing literal question-to-tag translation. Unlike prior autonomous taxonomy-generation systems, our approach (i) takes the per-occupation preference-tag catalog (not the hierarchy of occupations) as the unit of optimization, (ii) generates each occupation's tags independently in parallel, (iii) evaluates with marketplace-grounded critics contributing weighted penalties (no hard vetoes) to an adjusted score, and (iv) treats legacy-Q\&A-to-modern-tag migration as a distinct pipeline component rather than a downstream side effect. We report deployment results from a 14-day post-launch production cohort (1{,}840 enrolled pros, 9.3M filter evaluations) that surfaces a concrete catalog-hygiene gap that a global-build approach would have masked. 

\end{abstract}

\keywords{LLM autoresearch, taxonomy generation, marketplace catalog, multi-critic evaluation, LLM-as-judge, deployed system}

\maketitle

\section{Introduction}
\label{sec:intro}
\begin{figure*}[t]
    \centering

    \begin{subfigure}[t]{0.48\textwidth}
        \centering
        \includegraphics[width=\linewidth]{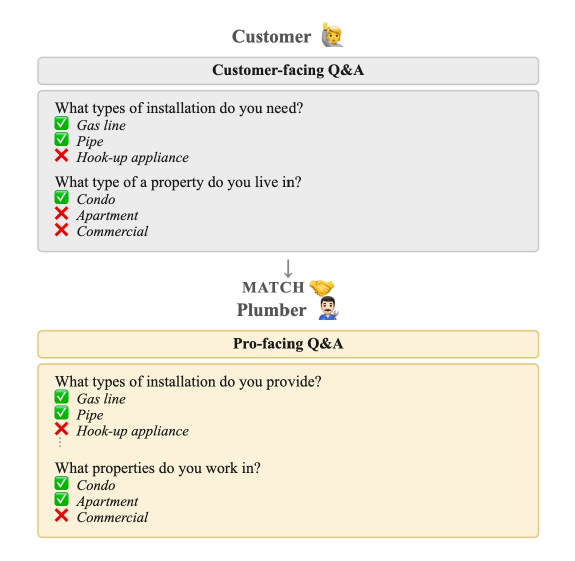}
        \caption{Legacy Q\&A matching}
        \label{fig:legacy-qa-matching}
    \end{subfigure}
    \hfill
    \begin{subfigure}[t]{0.48\textwidth}
        \centering
        \includegraphics[width=\linewidth,scale=0.5]{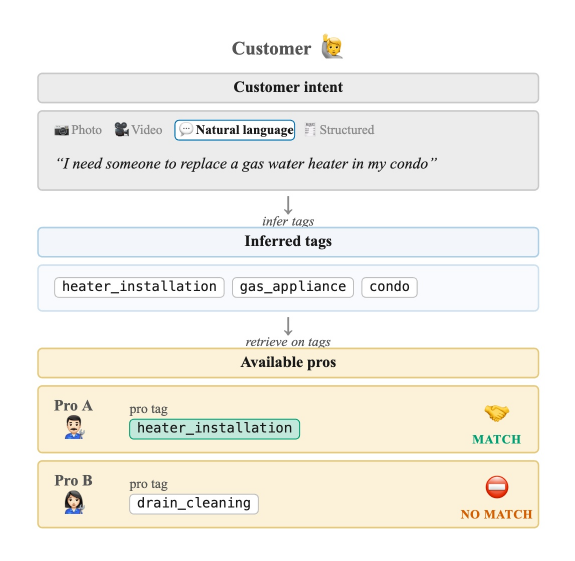}
        \caption{AI-native matching}
        \label{fig:nl-search-matching}
    \end{subfigure}

    \caption{
    Evolution from legacy request-form matching to AI-native matching: Legacy matching (\subref{fig:legacy-qa-matching}) relies on an implicit catalog embedded in occupation-specific Q\&A forms. AI-native matching (\subref{fig:nl-search-matching}) requires an explicit catalog that translates consumer intent and provider preferences into a shared representation for downstream marketplace systems.}
    \label{fig:matching-paradigms}
\end{figure*}

A marketplace catalog provides the shared representation through which consumer requests, provider preferences, and job attributes are interpreted. Matching, search, and pricing all depend on this common language. However, in service marketplaces that have historically relied on structured question-and-answer (Q\&A) forms, the catalog is often implicit, distributed across hundreds of category-specific schemas rather than represented as a unified structure. When the marketplace transitions to AI-native probabilistic matching (Figure \ref{fig:matching-paradigms}), this implicit catalog becomes a liability. Matching models require structured signals, pricing models require job attributes that transfer across categories, and marketplace participants need a coherent interface for expressing preferences without navigating hundreds of independent schemas. 

We frame this problem as a catalog \emph{reconstruction}, not \emph{creation}, since the catalog is already implicitly defined in the legacy system, albeit distributed, through Q\&A pairs. The task is to infer the latent attributes encoded across Q\&A pairs and reorganize them into an explicit representation suitable for AI-native marketplace interactions. Because the implicit legacy catalog is organized around distinct service domains, each with its own vocabulary, preference structure, and matching semantics, we decompose catalog reconstruction into a collection of smaller \emph{autoresearch} problems~\cite{karpathy2026autoresearch} rather than a single global taxonomy-generation task. 
%We instantiate this decomposition by treating each service domain as an independent autoresearch problem with its own propose-evaluate-keep loop. %
This design contrasts with recent hierarchical taxonomy-generation systems, which optimize coherent occupation hierarchies~\cite{li2025climb} or corpus-level research taxonomies~\cite{kargupta2025taxoadapt}. Our goal is not to build a single hierarchy over all domains, but to reconstruct preference catalogs whose semantics remain meaningful within each service domain in a two-sided marketplace.

The key contributions of this paper are as follows:
\begin{enumerate}
\item 
\textbf{Catalogs for marketplaces - }  We study a catalog- reconstruction problem distinct from occupation-hierarchy construction~\cite{li2025climb}, text-label taxonomy induction~\cite{wan2024tntllm}, research-corpus taxonomy adaptation~\cite{kargupta2025taxoadapt}, and product-schema modeling~\cite{amazon2025attributeforge}. The generated unit is a provider-facing \emph{preference tag}: a primitive that providers toggle with in production and that the downstream marketplace systems use.
\item
\textbf{Independent autoresearch loops - }
We decompose reconstruction into independent autoresearch jobs, one per service occupation, each with its own \emph{propose-evaluate-keep} loop. This avoids forcing all service domains into a single global hierarchy and limits semantic interference between occupations with superficially similar but operationally distinct attributes.
\item
\textbf{Marketplace-grounded automated evaluation - } We combine a recalibrated multi-rubric LLM-as-judge ~\cite{zheng2023judge} with a seven-persona critic panel that contributes weighted penalties to the generated catalog. This aligns evaluation with marketplace objectives rather than generic measures of taxonomy quality.
\item
\textbf{Legacy Q\&A parity mapping - }
We introduce a parity-mapping stage that maps the reconstructed catalog back to legacy request-form Q\&A pairs. This stage measures whether the new catalog covers the operational distinctions encoded in the legacy system and produces a human-reviewable artifact for validating migration from structured forms to provider-preference tags.
\end{enumerate}

\begin{comment}
\added{Success for this system means the marketplace can retire its legacy request forms without losing the preference structure they encoded: per occupation, a catalog that scores well under marketplace-grounded evaluation, passes human review, and behaves attributably once deployed. The strongest single piece of evidence is the deployment finding of \S\ref{sec:deployment}: when $40.66\%$ of production filter evaluations failed on a missing canonical tag, per-occupation independence made the failure attributable to deployment-side seeding rather than to the catalog.}
\end{comment}
Success for this system means the marketplace can retire its legacy request forms
without losing the preference structure they encoded. The most direct evidence is
reconstruction fidelity. Across the reconstructed catalog, $73.3\%$ of legacy answers map directly to a tag in the regenerated catalog; most of
the remainder are intake-only answers (sizes, ranges, ``flexible'') that a preference
schema should not carry, excluded as acceptable non-tags ($20.8\%$ of all answers). Only $6.0\%$ of legacy answers are \emph{regrettable} misses, real preferences left uncovered. Measured against the preferences that should map, pooled coverage is
$92.5\%$, the per-occupation median is $100\%$ (mean $97.8\%$). What the catalog drops is, by construction, intake detail; the screening preferences pros act on are almost entirely preserved. Together with production scale (132 occupations live, new ones onboarded on demand within hours) and mandatory human sign-off before deployment, this fidelity is what makes retiring the request forms defensible.

The propose-evaluate-keep loop builds on prior work in iterative LLM refinement, principle-guided critique, and autoresearch systems~\cite{madaan2023selfrefine,bai2022constitutional,karpathy2026autoresearch}; here, we operationalize it for production catalog reconstruction in a two-sided service marketplace. We report results from a production deployment spanning 132 service occupations as of April 2026, with on-demand onboarding of new occupations as the marketplace expands. The remainder of the paper describes the marketplace setting ($\S$\ref{sec:context}), other related works ($\S$\ref{sec:related}), the autoresearch system used for catalog reconstruction ($\S$\ref{sec:system_design}), its deployment in production ($\S$\ref{sec:deployment_in_prod}), and the empirical findings ($\S$\ref{sec:eval}) that emerge from operating the system at scale.

\section{Background}
\label{sec:context}
In this section, we describe the marketplace setting and the production context in which the system is deployed ($\S$\ref{subsec:marketplace_context}), as well as the terminology needed to understand the catalog-reconstruction problem ($\S$\ref{subsec:terminology}). 
\subsection{Marketplace context}
\label{subsec:marketplace_context}
The deploying organization is a major U.S. online services marketplace that connects consumers with local service professionals (\emph{pros}) across home, wellness, and event categories, including plumbing, house cleaning, photography, landscaping, and roughly 130 others. Pros specify targeting preferences that determine which consumer leads they are eligible to receive and purchase. The system described in this paper is part of a 2026 pro-side marketplace redesign that replaces rigid binary targeting defined through structured Q\&A with probabilistic matching driven by a structured provider-side preference taxonomy~\cite{einav2016peer}. 

\subsection{Terminology}
\label{subsec:terminology}

The legacy taxonomy is organized as a hierarchy of occupations, categories, and category-specific request-form Q\&A schemas. The reconstructed catalog contains two tag classes. 
\begin{enumerate}
\item \emph{Canonical tags} determine eligibility: a provider must possess the canonical tag associated with a category for the strict-match filter to consider that provider eligible for leads in that category; 
\item \emph{Specialty preference tags} allow providers to refine which leads they wish to receive within categories for which they are already eligible 
\end{enumerate}

Section~\ref{sec:eval} reports two production findings that depend on this distinction: a catalog-hygiene gap in which deprecated canonical tags were still emitted by request-time enrichment, and a deployment-side seeding gap in which $40.66\%$ of \emph{filter evaluations} failed because the canonical tag associated with the consumer request was not present on the provider profile. A \emph{filter evaluation} is a single matchmaker-side check, for one candidate pro and one incoming consumer request, to determine whether the pro's profile contains the request's required canonical tag.

\section{Related Work}
\label{sec:related}
Our work sits at the intersection of automated taxonomy construction, agentic schema generation, autoresearch systems, prompt optimization, and LLM-based evaluation.
\paragraph{Hierarchical taxonomy construction.} CLIMB~\cite{li2025climb} built occupation hierarchies using a global semantic clustering to distill core occupations, followed by a reflection-based multi-agent system to iteratively build a coherent hierarchy. Our work differs in both the unit of generation and the reconstruction objective: CLIMB generated occupations within a hierarchy, whereas we reconstruct provider-preference tags within an occupation. Furthermore, our catalogs are generated independently for each occupation and must support migration from legacy request-form Q\&A systems through parity mapping. TnT-LLM~\cite{wan2024tntllm} used LLMs to induce and iteratively refine label taxonomies from unstructured text, while TaxoAdapt~\cite{kargupta2025taxoadapt} dynamically adapted multidimensional taxonomies to evolving scientific corpora through iterative hierarchical classification. Both systems inform the generation-and-refinement paradigm we adopt. However, their objective is taxonomy induction for corpus organization and classification, whereas ours is catalog reconstruction for marketplace decision-making.

\paragraph{Schema and catalog generation.}
AttributeForge~\cite{amazon2025attributeforge} automated end-to-end product-schema modeling using a large collection of specialized LLM agents, together with automated evaluation and repair. It is the closest prior work to ours in terms of production-scale catalog reconstruction. The key distinction is the object being modeled: AttributeForge generated product attributes for e-commerce catalogs, whereas we reconstruct provider-preference catalogs for service marketplaces, where matching depends on both provider capabilities and provider screening preferences.

\begin{comment}

\paragraph{Multi-agent product-schema modeling.} AttributeForge~\cite{amazon2025attributeforge} uses 43 LLM agents to model product schemas for e-commerce at scale. The closest prior art on multi-agent catalog construction. We differentiate on product-vs-service framing (a service catalog has to model both \emph{what the pro does} and \emph{what they screen out}, while a product catalog primarily models what the product \emph{is}) and on pro-side vs product-side optimization target.

\paragraph{Autoresearch / self-refinement loops.} Karpathy's autoresearch~\cite{karpathy2026autoresearch} popularized the three-file loop (fixed eval, editable artifact, agent instructions). Self-Refine~\cite{madaan2023selfrefine}, Constitutional AI~\cite{bai2022constitutional}, multi-agent debate~\cite{du2024multiagent}, and GEPA~\cite{agrawal2025gepa} are direct ancestors. We use a naive LLM proposal as the search backend; GEPA-style reflective optimization is planned as an upgrade.
\end{comment}

\paragraph{Autoresearch, self-refinement, and prompt optimization.}
Our "propose, evaluate, keep" loop draws on a broader family of iterative LLM optimization methods, including Self-Refine~\cite{madaan2023selfrefine}, Constitutional AI~\cite{bai2022constitutional}, multi-agent debate~\cite{du2024multiagent}, GEPA~\cite{agrawal2025gepa}, and prompt-optimization frameworks such as DSPy and MIPROv2~\cite{opsahlong2024miprov2}. Like DSPy, we treat prompts as optimizable artifacts and evaluation as the search signal. However, rather than optimizing a single prompt against a fixed benchmark or held-out metric, our framework optimizes occupation-specific catalog-generation prompts whose outputs are structured marketplace catalogs. 

\begin{comment}
\paragraph{Programmatic prompt optimization.} The closest prior art on \emph{automated prompt optimization} is the DSPy program-of-prompts framework and its bundled optimizer MIPROv2~\cite{opsahlong2024miprov2}, which jointly optimizes instructions and few-shot demonstrations for multi-stage LLM programs against a held-out metric. We share DSPy's commitment to treating the prompt as the optimized artifact and a fixed evaluator as the search signal. We differ on two dimensions: (i) MIPROv2 optimizes a single prompt-program in a domain where the evaluator is a numeric metric on a held-out task, while our loop optimizes a per-occupation prompt against a composite of an LLM-as-judge and a critic panel grounded in marketplace-side data; (ii) per-occupation independence gives 132 small optimization problems running in parallel, rather than a single program optimized over a global benchmark. GEPA~\cite{agrawal2025gepa} sits closer to our setting in that it uses execution traces as the reflection signal, and is explicitly the planned upgrade to our search backend once the critic-panel output is structured enough to serve as a reflection signal.
\end{comment}
\begin{figure}[t]
\centering
% S2 publication diagram 1 — Per-occupation autoresearch loop (conceptual overview)
% Single-column layout: vertical top-to-bottom flow so it fits \columnwidth.
% Companion to Algorithm~\ref{alg:autoresearch}; deliberately formula-free so the two
% artifacts complement rather than duplicate each other.
% Requires: \usetikzlibrary{positioning,fit,arrows.meta}
% Use inside a single-column figure (NOT figure*) env.
\begin{tikzpicture}[
    >={Stealth[length=2mm]},
    every node/.style={font=\small},
    box/.style={draw, very thin, rectangle, rounded corners=1pt, fill=black!4,
                minimum height=7mm, minimum width=26mm, align=center, inner sep=2pt},
    data/.style={draw, very thin, rectangle, fill=white,
                 minimum height=6.5mm, align=center, inner sep=3pt},
    critic/.style={draw, very thin, rectangle, rounded corners=1pt, fill=black!4,
                  font=\scriptsize, minimum height=5mm, minimum width=18mm,
                  align=center, inner sep=1pt},
    decision/.style={draw, semithick, diamond, aspect=1.6, fill=black!12,
                     align=center, inner sep=1pt, font=\footnotesize},
    annot/.style={font=\scriptsize\itshape, align=center},
]
  % --- Input: legacy RF data at top, feeding the Generator
  \node[data] (legacy) {Legacy RF data $D_o$};
  \node[box, below=8mm of legacy] (gen) {Generator};
  \draw[->] (legacy) -- (gen);
  % --- E1 judge
  \node[box, below=6mm of gen] (e1) {Judge\,(score /15)};
  \draw[->] (gen) -- (e1);
  % --- Critic panel: 7 personas in a compact 2-column grid
  \node[critic, below=13mm of e1, xshift=-10mm] (c1) {PM};
  \node[critic, right=2mm of c1] (c2) {Pro};
  \node[critic, below=1.5mm of c1] (c3) {Taxonomy};
  \node[critic, right=2mm of c3] (c4) {Adversarial};
  \node[critic, below=1.5mm of c3] (c5) {Reasoning};
  \node[critic, right=2mm of c5] (c6) {Coherence};
  \node[critic, below=1.5mm of c5] (c7) {Parity};
  \node[fit=(c1)(c2)(c7), draw, very thin, dashed, inner sep=3pt] (panel) {};
  \draw[->] (e1.south) -- (panel.north);
  % Panel title sits on the connector (white fill masks the line) -> reads as a label.
  \node[annot, fill=white, inner sep=1pt, above=2mm of panel.north]
        {7-persona critic\\panel (parallel)};
  % --- Aggregate -> decision
  \node[box, below=6mm of panel] (agg) {Aggregate penalties\\(set-level + critic)};
  \draw[->] (panel.south) -- (agg.north);
  \node[decision, below=6mm of agg] (keep) {Improved?};
  \draw[->] (agg) -- (keep);
  % --- Editor on the left; feedback loop runs up the left margin into the Generator
  \node[box, left=9mm of keep] (editor) {Editor:\\revise prompt};
  \draw[->] (keep.west) -- node[above, font=\scriptsize] {iterate} (editor.east);
  \draw[->] (editor.west) -- ++(-3mm,0) |- (gen.west);
  % --- On plateau / budget: emit best catalog, then parity mapping
  \node[data, below=6mm of keep] (best) {Best catalog $T^\star$};
  \draw[->, dashed] (keep.south) -- node[right, font=\scriptsize] {on plateau} (best.north);
  \node[box, below=5mm of best] (parity) {Parity mapping\\(tags $\leftrightarrow$ RF Q\&A)};
  \draw[->] (best) -- (parity);
  \node[annot, below=1mm of parity] {coverage + human QA};
\end{tikzpicture}
\caption{Per-occupation autoresearch loop (conceptual overview; see Algorithm~\ref{alg:autoresearch} for the precise procedure). The occupation's legacy RF data $D_o$ and current best prompt $p_o$ feed a Generator that
produces a candidate tag set. Each candidate is scored by the $\mathcal{E}$ six-rubric LLM-as-judge; the seven-persona critic panel then runs in parallel, and its feedback is aggregated. The gate decides whether the editor's single proposed prompt edit improves on the current best and is kept. On a plateau or budget exhaustion, the best catalog $T^\star$ is handed to the post-loop parity mapping,
which links catalog tags to legacy RF Q\&A (many-to-many) and produces the artifact humans review.}
\label{fig:autoresearch-loop}
\end{figure}
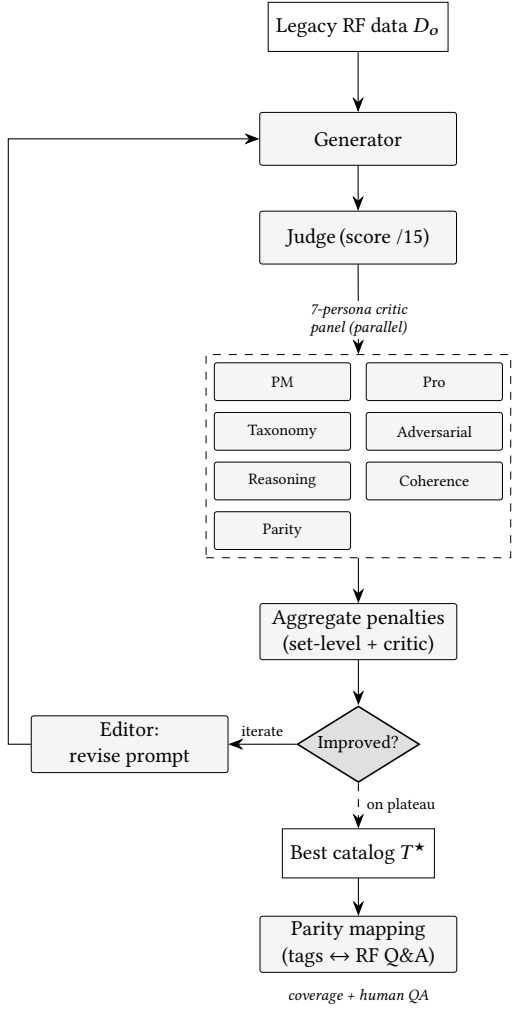
\paragraph{LLM-based evaluation and abstraction.}
Our system draws on recent work in both LLM-based evaluation and abstraction. The six-rubric evaluation framework builds on LLM-as-judge methodologies~\cite{zheng2023judge} and critique-based evaluation~\cite{husain2024critique}, adapting them to catalog reconstruction through marketplace-specific criteria derived from production review. Several rubrics were introduced to capture failure modes that generic taxonomy metrics overlook, such as tags that are semantically coherent but insufficiently specific or interpretable for provider-facing use. We also draw inspiration from Step-Back Prompting~\cite{zheng2024stepback}. Rather than translating legacy request-form questions directly into tags, we first infer the underlying job attribute each question was intended to capture and use that abstraction as input to catalog generation.

\section{System Design}
We formulate catalog reconstruction as an iterative propose-evaluate-keep problem ($\S$\ref{sec:loop}). Candidate catalogs are generated independently for each service domain, evaluated using a rubric-based LLM judge ($\S$\ref{sec:e1}), adjusted through a multi-persona critic panel ($\S$\ref{sec:critics}), and validated against legacy request-form semantics through parity mapping ($\S$\ref{sec:parity_map}).
\label{sec:system_design}

\subsection{Per-Occupation Autoresearch Loop}
\label{sec:loop}
Catalog reconstruction is a cold-start problem: initially there is no per-occupation generation prompt, so every occupation $o \in O$ ($|O| = 132$ in the current production
deployment) is seeded with the same baseline prompt $p^{(0)}$, i.e. $p_o^\star \!\gets\! p^{(0)}$. We run Algorithm~\ref{alg:autoresearch} (illustrated in
Figure~\ref{fig:autoresearch-loop}) independently per occupation. 

At iteration $0$, the generator $\mathcal{G}$ applies the baseline prompt to the occupation's legacy data $D_o$ ( RF Q\&A schema for each category, \% share of recently-active pros who have currently enrolled a (q, a) pair as a preference) to produce a candidate tag set $T$. The six-rubric LLM-as-judge scores the set, $s = \mathcal{E}(T) \in [0, 15]$, and the seven-persona critic panel adds weighted penalties $\pi_k$, giving the composite score $a = s - \sum_k w_k \pi_k(T)$. This becomes the initial best
$(p_o^\star, T^\star, s^\star, a^\star)$. The loop runs a cross-family model stack: the generator $\mathcal{G}$ and judge $\mathcal{E}$ on one family (GPT-5-4 / GPT-5-4-mini), the seven critics, editor, and parity mapper on another (Claude Sonnet 4.6), so the models that propose and score do not share biases with the models that critique and mutate (\S\ref{sec:e1}, \S\ref{sec:discussion}).

Each subsequent iteration proposes and tests a single prompt edit. The editor $\mathcal{R}$ makes one targeted change to the current best prompt $p_o^\star$, focused
on a single weak quality dimension: it ranks the six $\mathcal{E}$ rubrics together with an injected parity-coverage dimension by mean score and selects the lowest-scoring dimension not targeted in the previous two iterations, avoiding fixation on any one rubric. The editor conditions on the weakest tag examples in that dimension, the score trends across iterations, and the log of previously accepted and rejected edits. The revised prompt is regenerated and rescored; as a per-iteration cost saver, the critic panel is invoked only when the candidate's $\mathcal{E}$ score is not already below the best ($s \ge s^\star$), since a lower-$\mathcal{E}$ candidate is very unlikely to overcome the best composite score. The edit is accepted,
thus becoming the new best, \emph{iff} its composite score improves, $a > a^\star$; otherwise the prompt reverts to $p_o^\star$. The loop runs for a fixed budget of $B$ iterations (default $5$). The supplementary material traces one iteration end-to-end on the \emph{Accounting} occupation. 

On termination, the parity mapping $\mathcal{P}(T^\star, Q_o)$ links the tags of the best catalog $T^\star$ to the legacy RF Q\&A $Q_o$ as a many-to-many relation $M^\star \subseteq T^\star \times Q_o$, leaving residual sets $U_Q^\star$ (RF Q\&A covered by no tag---typically intake-specific size/range/flexibility questions) and $U_T^\star$ (tags covered by no Q\&A). We emit $M^\star$ as a JSON mapping and a Google Sheet for human review before deployment.

\begin{algorithm}[t]
\caption{Occupation-Local Autoresearch}
\label{alg:autoresearch}
\begin{algorithmic}[1]
\Require legacy data $D_o$ with RF Q\&A $Q_o$, shared seed prompt $p^{(0)}$, budget $B$, critic weights $w$
\Ensure best prompt $p_o^\star$, best catalog $T^\star$, parity relation $M^\star$, unmapped sets $U_Q^\star, U_T^\star$
\State $p_o^\star \!\gets\! p^{(0)},\;\; T^\star \!\gets\! \mathcal{G}(p_o^\star, D_o),\;\; a^\star \!\gets\! \mathcal{E}(T^\star) - \rho(T^\star) - {\textstyle\sum_k} w_k\pi_k(T^\star)$
\For{$t = 1$ to $B$}
    \State $p_o \gets \mathcal{R}(p_o^\star),\;\; T \gets \mathcal{G}(p_o, D_o),\;\; \hat{a} \gets \mathcal{E}(T) - \rho(T)$ \Comment{pre-critic score; $\rho$ = set-level penalty (parity coverage + proliferation)}
    \If{$\hat{a} > a^\star$} \Comment{$\hat{a} \ge a$, so a candidate with $\hat{a}\le a^\star$ provably cannot win; skip critics}
        \State $a \gets \hat{a} - {\textstyle\sum_k} w_k\pi_k(T)$
        \If{$a > a^\star$} \State $(p_o^\star, T^\star, a^\star) \gets (p_o, T, a)$ \EndIf
    \EndIf
\EndFor
\State $M^\star \gets \mathcal{P}(T^\star, Q_o) \subseteq T^\star \times Q_o$ \Comment{many-to-many tag\,$\leftrightarrow$\,Q\&A relation}
\State $U_Q^\star \gets Q_o \setminus \pi_Q(M^\star)$ \Comment{unmapped Q\&A}
\State $U_T^\star \gets T^\star \setminus \pi_T(M^\star)$ \Comment{tags with no covering Q\&A}
\State \Return $(p_o^\star, T^\star, M^\star, U_Q^\star, U_T^\star)$
\end{algorithmic}
\end{algorithm}

\subsection{Six-Rubric LLM-as-Judge}
\label{sec:e1}
The judge model $\mathcal{E}$ scores every tag ($ t\in T$) on six dimensions:

\begin{itemize}
    \item \textbf{Screening vs.\ Intake} (0--3): Does the tag describe a job type the pro might want \emph{more or less of} (good), or is it a customer-intake detail the pro should not screen on (bad)? E.g.\ ``Tax return preparation'' (screening) vs.\ ``Travel range: 15 miles'' (intake).
    \item \textbf{Tag Legibility} (0--3): Would a working pro understand this tag from the tag text alone? E.g.\ ``Appliance repair'' (clear) vs.\ the bare noun ``Appliances''
    \item \textbf{Preference Variance} (0--3): Do real pros split on this tag, or is everyone equally for-or-against (degenerate)? E.g.\ Accounting pros split on ``QuickBooks'' proficiency, but nearly all accountants serve the ``healthcare industry'' (degenerate).
    \item \textbf{Cross-Category Consistency} (0--2): When the same job-attribute appears in two categories within the occupation, do the tags match? E.g.\ the client-entity type tagged identically as ``S-Corp'' wherever it recurs across the occupation's categories, not ``S-Corp'' in one and ``S-Corporation'' in another.
    \item \textbf{Canonical Coverage} (0 or 2): Does every category in the occupation have a canonical tag covering it? E.g.\ ``Payroll services'' tag cleanly covers the \emph{Payroll Services} category, whereas a tag backed by no category scores 0
    \item \textbf{Information Loss} (0--2): Are the tag's sources traceable to specific legacy Q\&A answers, i.e.\ does it cite the legacy category and answers it consolidates? E.g.\ a tag citing \texttt{legacy\_category: Accounting} and the ``QuickBooks'' answer scores 2 vs.\ one with only a vague rationale ($0$).
\end{itemize}

$\mathcal{E} $ scores each tag independently on the six rubrics above, with all tags in $T$ scored in parallel; the per-tag composite is the sum of the six rubric scores (on a $/15$ scale). The set-level score is the mean per-tag composite reduced by small set-level deductions for tag proliferation and category-coverage gaps, $s = \mathcal{E}(T) \in [0, 15]$; the verdict mix summarizes the per-tag outcomes. The judge model is GPT-5-4-mini at temperature zero; the generation model is GPT-5-4. The seven-persona critic panel, the editor agent, and the final parity-mapping stage all run on Claude Sonnet 4.6, a different model family from the generator and the judge. The choice is deliberate: a same-family stack would risk the loop reinforcing biases shared across mutation and evaluation, and the cross-family arrangement is the strongest readily available defense at production scale. $\mathcal{E}$ was recalibrated against product-manager (PM) review: bare-noun tags like ``Appliances'' originally scored a legibility of $3$ but scored $\leq 1$ after recalibration, since a working appliance-repair pro reads ``Appliances'' as either trivially true or meaningless, not informative.

\subsection{Set-Level Penalty ($\rho$)}
\label{sec:set-penalty}
The $\mathcal{E}$ judge scores each tag in isolation, so a catalog of individually high-scoring tags can still be poor as a \emph{set}. $T$ can be bloated with near-duplicate siblings, or leave real pro preferences with no corresponding tag. The set-level penalty $\rho(T)$ corrects for this by deducting from the composite two failures the per-tag rubrics cannot see: \emph{tag proliferation} (an excess of sibling tags under a single parent, which fragments the catalog and the pro-facing UI) and \emph{coverage gaps} (legacy preferences in $Q_o$ which pros both enrolled and deselected recently that no tag captures, including whole categories left untagged; $\S$\ref{sec:parity_map}). Subtracting $\rho(T)$ before the critic penalties ($\S$\ref{sec:critics}), $a = s - \rho(T)$, makes the loop optimize for a coherent, well-covered catalog rather than a collection of locally excellent but globally redundant or incomplete tags.

\subsection{Seven-Persona Critic Panel}
\label{sec:critics}
After $\mathcal{E}$ scoring, the full tag set is reviewed by seven critic-personas running in parallel, each returning a verdict that is reduced to a weighted contribution to the composite score. Most contributions are penalties; the Reasoning critic can instead award a small bonus for strong rationales. The exact penalty formulas, caps, and bonus thresholds are tabulated in the supplementary material.\footnote{An earlier design used a hard veto from the Pro Critic. In early production runs, the veto dominated keep/discard decisions, and the loop stalled on occupations where one persona consistently rejected proposals that other critics rated favorably. The current design replaces the veto with a weighted penalty capped at $2.5$ on the composite-of-15 scale; low enough that a strong proposal can still be kept over Pro objections, high enough that consistent Pro disapproval pulls the adjusted score below the discard threshold. The cap and rate were set empirically; the panel module records the change as ``softened to weighted penalty, capped at 2.5.''} The critics are:

\begin{itemize}
    \item \textbf{PM critic}: Would a product manager (PM) find this tag actionable? Surfaces design-stage issues (mutual-exclusivity, awkward dimensionality); penalized only when rejections exceed a small free allowance.
    \item \textbf{Pro critic} (penalty capped at $2.5$): Would a working pro in this occupation find this useful or confusing? The cap prevents a single skeptical persona from dominating the verdict.
    \item \textbf{Taxonomy critic}: Are tags at consistent levels of abstraction, are unions broken into components, are subtypes grouped? Penalized as structural consistency falls below the threshold.
    \item \textbf{Adversarial critic}: Where could the tag set be gamed or misinterpreted to extract leads that the pro cannot deliver? Penalty scales with the assessed risk level and the number of critical failures.
    \item \textbf{Reasoning-Quality critic} (bonus-eligible): Does the generation explain \emph{why} each tag exists? Strong rationales (score $\geq 8$) earn a small bonus; weak ones incur a penalty.
    \item \textbf{Occupation-Coherence critic}: Does the tag set, taken as a whole, describe \emph{this occupation} rather than a generic union of services?
    \item \textbf{Parity critic}: Do the generated tags map to a given occupation's top-30 most-deselected legacy answers (i.e., answers active pros have explicitly opted out of), ranked by distinct-pro count over a 90-day window? Its penalty is small by design (at most $\sim\!1$ point); the quantitative coverage enforcement lives in the set-level penalty $\rho$ (\S\ref{sec:set-penalty}). When an occupation has no deselection signal, the critic returns a neutral score and contributes nothing. 
\end{itemize}    

\subsection{Editor Agent}
\label{sec:editor}
The editor $\mathcal{R}$ is the loop's mutation operator: at each iteration, it proposes exactly one targeted edit to the current best prompt $p_o^\star$ and returns the revised prompt together with a one-line description of the change. A key design point is that the editor is driven by the judge $\mathcal{E}$, not the critics: the critic penalties gate which prompts survive (through the composite $a$), whereas the editor decides \emph{what to change} from the $\mathcal{E}$ signal and the coverage signal alone.

\paragraph{Target-dimension selection.}
The editor ranks two complementary kinds of signals and targets the weakest. The first is \emph{per-tag quality}: the six $\mathcal{E}$ rubrics, each averaged across the tag set to yield one mean per rubric. The second is \emph{set-level coverage}: a single parity-coverage score for the catalog as a whole. Coverage cannot be expressed as a seventh rubric, since $\mathcal{E}$ scores the tags that exist, whereas coverage is a property of the legacy preferences that have \emph{no} tag, so there is nothing per-tag to score. It is therefore injected as a synthetic dimension, giving catalog incompleteness a seat in the ranking alongside tag-quality weaknesses. The coverage dimension is a \emph{score} (higher is better), $2\min(r_{+}, r_{-})\in[0,2]$, where $r_{+}, r_{-}\in[0,1]$ are the fractions of enrolled and deselected preferences the catalog covers; the $0$--$2$ scaling puts it on the same footing as the rubric means. This ranking score is used \emph{only} to select the target dimension and is distinct from the set-level penalty $\rho$ (\S\ref{sec:set-penalty}), which turns the same coverage signal into points subtracted from the composite. The editor then targets the lowest-scoring dimension that has not already been addressed in the previous N iterations (default N = 2), a constraint that prevents fixation on a single persistent weakness and forces progress across both quality and coverage rather than over-optimizing one axis.

\paragraph{Conditioning.}
For the chosen dimension, the editor is given the tags scoring weakest on it ($\mathcal{E}$ rationales with a score $\leq 1$, or the specific unmapped preferences when parity-coverage is targeted), the dimension's score trend across iterations (improving / stable / worsening), and a log of prior edits annotated with their kept/discarded outcomes. It is instructed to propose exactly one change, to try a fundamentally different approach when a dimension keeps worsening despite edits targeting it, and never to repeat a previously discarded change. The revised prompt $p_o^{(t+1)}$ is then regenerated and rescored (\S\ref{sec:loop}), and kept only if its composite score improves.

\subsection{Parity mapping}
\label{sec:parity_map}
Once the loop terminates with the best catalog $T^\star$, a final LLM pass migrates the occupation's legacy RF Q\&A $Q_o$ onto the new tags, producing the many-to-many relation $M^\star \subseteq T^\star \times Q_o$ returned by Algorithm~\ref{alg:autoresearch}. The autoresearch design ensures that the generated catalog covers a good fraction of legacy taxonomies (explicitly in $\S$\ref{sec:set-penalty}, parity critic in $\S$\ref{sec:critics}, and set level coverage in $\S$\ref{sec:editor}). $M^\star$, on the other hand, is \emph{explicit and auditable}: it records exactly which legacy preferences map to which tags and which remain uncovered, producing the human-reviewable artifact used to sign off the catalog before deployment. For each legacy answer, the mapper finds \emph{all} tags that cover the same real-world concept; a single answer may map to several tags (for example, a specific tag and the broader tag above it), which is why $M^\star$ is a many-to-many relation rather than a function. 
\begin{comment}
\shishir{added: the intro's fidelity numbers need a derivation in the body --- Kartik, verify these match your analysis}
\end{comment}
Aggregated across the 132-occupation production catalog, $M^\star$ is also what yields the reconstruction-fidelity numbers quoted in \S\ref{sec:intro}: $73.3\%$ of legacy answers map directly to a tag in the regenerated catalog; $20.8\%$ are intake-only answers (sizes, ranges, ``flexible'') excluded as acceptable non-tags; the remaining $6.0\%$ are regrettable misses, real preferences left uncovered. Measured against the answers that should map, pooled coverage is $92.5\%$, with a per-occupation median of $100\%$ (mean $97.8\%$).

\section{Deployment}
\label{sec:deployment_in_prod}
The reconstructed catalog has been running in production since April 2026. This section describes how catalogs were generated at scale across the occupation set and how each catalog was staged for human review before going live.

\paragraph{Catalog Generation}
Each occupation is reconstructed independently by the per-occupation loop of \S\ref{sec:loop}: starting from the shared baseline prompt, the generator, judge, critic panel, and editor iterate for the fixed budget, after which the parity mapping produces the occupation's best catalog $T^\star$, its tag$\leftrightarrow$Q\&A relation $M^\star$, and the residual sets $U_Q^\star, U_T^\star$. Because the loop is occupation-local, the 132 occupations are processed independently and in parallel. Every run persists a full audit trail: a per-occupation iteration log recording each kept change with its rationale and target dimension, a snapshot of the best tag set and its scores at each iteration, and the parity-mapping outcome. These artifacts are both retained per run and streamed to a shared reviewer workspace as iterations are kept (\S\ref{sec:human-qa-at-scale}), so the catalog and its rationale are available for inspection while a run is still in progress.

\paragraph{Human QA at Scale}
\label{sec:human-qa-at-scale}
No catalog is deployed without human sign-off, so each generated catalog is seeded into a shared Google Sheet for review. Results were uploaded live during a run into managed per-occupation tabs: \emph{Best} (the current tag set), \emph{Iteration Log} (every kept change with its rationale), \emph{Tricky Parity Cases} (ambiguous legacy mappings flagged for attention), and \emph{Comment Archive} (resolved reviewer comments). The parity tab includes a Verdict dropdown (Keep/Update/Delete) with conditional formatting, and reviewers can inline-edit tag names. Review is not a one-way gate: unresolved reviewer comments are polled from the sheet (via the Drive API) and written to a per-occupation feedback file, which the generator and editor read on the next run. Human feedback, such as renamed tags, deletions, and free-text objections, therefore feeds directly back into the automated loop, closing the gap between manual review and regeneration.

\paragraph{Scale and Coverage}

The system has been live across 132 occupations. New occupations are onboarded on demand from a single configuration entry: a per-occupation config file maps each occupation slug to its data-warehouse table names and primary keys, and adding an occupation requires only that entry. \emph{Automotive Detailing}, for example, was added and taken end-to-end: baseline generation, autoresearch loop, parity mapping, and human QA review, within five hours on a single day in May 2026.

\section{Evaluation and Operational Lessons}
\label{sec:eval}

\subsection{Catalog Quality vs.\ Deployment Outcomes}
\label{sec:catalog-vs-deployment}

The autoresearch loop produces a catalog whose quality is measured in one space (the $\mathcal{E}$ composite plus the seven critic penalties 
\begin{comment}
on cohorts (B)--(D), \S\ref{sec:cohorts}
\end{comment}
), and the deployment is measured in another (filter-evaluation outcomes, occupation concentration). 
\begin{comment}
hygiene gaps observed against the live event log on cohort (E)
\end{comment}
Conflating the two muddles attribution: a missing-canonical-tag filter rate of $40.66\%$ in production is not a verdict on the catalog, it is a verdict on the deployment-side seeding step that was supposed to attach canonical tags to existing pro profiles. Likewise, the loop-internal critic penalties say nothing about whether request-time enrichment actually emits the right tags in production. We separate the two strands explicitly in what follows.

\subsection{Catalog-Quality Metrics}
\label{sec:catalog-quality}

\paragraph{$\mathcal{E}$ composite distributions.} Across the 132-occupation production runs, the per-tag $\mathcal{E}$ composite (out of 15) lands in a relatively narrow band on the best kept catalog per occupation: the bulk of tags score in the 11--14 range, with low-end outliers concentrated on Canonical Coverage (the binary $0$-or-$2$ rubric) for occupations whose legacy categories do not have an obvious single-tag canonical mapping. 
\begin{comment}
The companion analysis notebook reproduces these distributions from the live tracking sheet's per-occupation \emph{Best} tabs.
\end{comment}

\paragraph{Critic-panel score distributions.} 
\begin{comment}
The same notebook extracts critic-panel scores per kept iteration across all occupations. 
\end{comment}
As designed, the Pro critic accumulates the largest aggregate penalty mass (capped at $2.5$ per iteration; supplementary material); the Adversarial critic contributes a sparse but heavy-tailed distribution dominated by the \texttt{critical} risk tier; and the Reasoning bonus fires on roughly the iterations where the editor's most recent change was a rationale upgrade.

\paragraph{Cross-occupation tag-name drift.} Because each occupation's catalog is generated in isolation, two occupations whose tags should share a name could drift apart. We measure this on a 14-occupation, 189-tag cross-section at two layers:

(i) Pro-facing tag strings (the names shown to pros, e.g., Cabinet installation) all 189 are unique to their occupation. Occupation-local vocabulary is preserved as intended: chess tutoring emits Adult chess tutoring, not a generic Adult that would also fit physical therapy. This is an exact-string check; semantic near-duplicates like Installation vs Installations are a within-occupation Taxonomy-critic concern, not a cross-occupation one.

(ii) Structural category prefixes (the part before the colon in each tag's structured form, e.g., Work type in Work type: Cabinet installation — the axis the matchmaker and pricing model filter on) 12 distinct prefixes across the cohort after case-folding, distributed as shown in Table~\ref{tab:kv-drift}. Four are widely shared: Capability and Work type in 9/14 occupations, Deliverable and Scope in 7/14, confirming partial standardization emerges without a global hierarchy. Two residual drifts are both targets for the next autoresearch wave: (a) eight singleton prefixes, three of which are clearly the same concept under different names (Client preference, Client type, Customer type); and (b) case inconsistency on shared prefixes (e.g., Deliverable vs DELIVERABLE).

\begin{table}[t]
\caption{Cross-occupation structural-prefix (KV-key) drift on the $n=14$ occupation cross-section (189 tags), case-folded. The four shared keys confirm partial standardization without a global hierarchy; the eight singletons and case variants are the residual drift targeted by the next autoresearch wave. }
\label{tab:kv-drift}
\small
\begin{tabular}{lr}
\toprule
Structural prefix (KV-key) & Occupations \\
\midrule
\texttt{Capability} & 9 / 14 \\
\texttt{Work type} & 9 / 14 \\
\texttt{Deliverable} & 7 / 14 \\
\texttt{Scope} & 7 / 14 \\
8 singleton prefixes (each) & 1 / 14 \\
\bottomrule
\end{tabular}
\end{table}

\paragraph{Planned dual-evaluator validation.} A complementary validation we plan (eval-of-eval) compares the production $\mathcal{E}$ judge against an independently-developed external evaluator (different judge model, different rubric structure, different thresholds) on the same per-occupation tag set, structured around verdict-agreement rate, per-rubric Cohen's $\kappa$ on the six shared dimensions, and a critic-based qualitative analysis of disagreement tags.

\paragraph{Critic ablation.} A single-occupation Parity-removed critic ablation on \emph{Accounting} (supplementary material) re-aggregates the production iteration log under a six-critic configuration. On the observed Accounting run, the Parity critic returned \texttt{parity\_score} $= 7/10$ in both kept iterations — above the panel's penalty thresholds — and therefore contributed zero to the critic penalty. Removing Parity from the panel leaves the adjusted-score trajectory and the final best tag set unchanged on this occupation. We read this as evidence that, on occupations where parity-readiness is comfortably above the penalty floor, the Parity critic's seat is insurance against the lower-readiness tail rather than an active per-iteration contributor; characterizing Parity's marginal contribution across the 132-occupation cohort is the next ablation-program step.

\subsection{Deployment-Side Metrics}
\label{sec:deployment}

The 14-day production post-launch cohort study (
\begin{comment}
cohort (E), \S\ref{sec:cohorts}: 
\end{comment}
1{,}840 MVP-enrolled pros, 9{,}324{,}798 filter evaluations over 2026-04-28 to 2026-05-12) characterizes the catalog \emph{as deployed}, distinct from the catalog as generated.

\paragraph{Catalog hygiene.} Three catalog tags marked \texttt{status = 2} (deprecated) were still being emitted by request-time enrichment as \emph{canonical category tags}, driving roughly $486{,}000$ filter events in the window. The implicated tags---\emph{Heavy lifting}, \emph{Bathroom remodel}, \emph{Wallpaper installation or repair}---came from a request-side reverse-mapping table that was never reconciled with the catalog's active-status flag. The autoresearch loop did not surface this finding directly: it produced a per-occupation catalog whose deployment, monitored at the filter-evaluation event level, exposed the request-side / pro-side mismatch. The \emph{surfacing} came from per-event deployment monitoring---any per-event monitoring pipeline against a hierarchical or per-occupation build could have detected the same SQL signal. What per-occupation independence specifically contributes is unambiguous \emph{attribution}: in a unified hierarchy ``Heavy lifting'' has a defensible existence as a Moving-and-Lifting sub-tag, so the deprecation flag is one signal among many and the diagnosis is ``data-quality nuisance''; under per-occupation independence the deprecated tag has no occupation that claims it as canonical, so the request-side emission is unambiguously a request-side / pro-side mismatch and the fix path is mechanical (re-activate the tag or remove from the reverse mapping). The contribution is the diagnostic clarity, not the diagnostic visibility.

\paragraph{Filter outcome distribution.} For the same cohort, the strict-match filter outcome distribution is shown in Table~\ref{tab:filter-outcomes}. The dominant failure mode---missing canonical category tag, $40.66\%$ of all MVP evaluations---is a deployment-stage gap, not a catalog-quality gap: the catalog tag exists, request-side enrichment fires it, but the pro's profile was never seeded with it. This motivates the deployment-side intervention (auto-attach canonical category tags at MVP onboarding) and validates the Parity critic's instinct that catalog completeness without deployment-side wiring is not production coverage.

\begin{table*}[t]
\caption{Strict-match filter outcome distribution across 9.3M MVP filter evaluations in the 14-day post-launch window. The dominant failure mode---missing canonical category tag---accounts for 40.66\% of all MVP evaluations and is, by itself, larger than the sum of all other filter reasons combined.}
\label{tab:filter-outcomes}
\begin{tabular}{lrr}
\toprule
Outcome & Filter evaluations & \% of MVP evals \\
\midrule
PASSED & 3{,}898{,}008 & 41.80\% \\
FILTERED: missing canonical category tag & 3{,}791{,}760 & 40.66\% \\
FILTERED: missing hybrid-preference data & 743{,}341 & 7.97\% \\
FILTERED: no canonical tag, no category-level targeting & 568{,}254 & 6.09\% \\
FILTERED: limitation on RF preference tag & 306{,}531 & 3.29\% \\
FILTERED: no targeted occupation match & 14{,}740 & 0.16\% \\
FILTERED: limitation on canonical preference tag & 2{,}164 & 0.02\% \\
\midrule
\textbf{Total} & \textbf{9{,}324{,}798} & \textbf{100.00\%} \\
\bottomrule
\end{tabular}
\end{table*}

\paragraph{Occupation concentration.} The same cohort surfaced a strong concentration of filter events: \emph{Handyman} alone accounted for roughly $70\%$ of MVP filter events ($3.77$M of $5.43$M filtered evaluations). Four other occupations had structurally high trigger rates: \emph{Roofing/siding} ($73.5\%$), \emph{Home technology} ($72.8\%$), \emph{Doors/windows} ($67.0\%$), \emph{Appliances} ($66.5\%$). This suggests the canonical tag taxonomy for these occupations does not match how pros describe their work. These are concrete targets for the next autoresearch wave, identifiable because per-occupation results are not averaged into a global score.

\paragraph{Directional post-launch signals (extended MVP window).} We did an MVP analysis at $8$--$10$ weeks post-launch (June--July 2026) and opened up enrollment from the $1{,}840$-pro cohort above to $\sim$$3{,}900$ pros. We got directional, though not causal, evidence consistent with the generated-vs-deployed attribution of \S\ref{sec:catalog-vs-deployment}. (i) About $88\%$ of enrolled pros remained enrolled, with attrition concentrated among high-volume providers and attributed to price and lead-mix economics, not catalog semantics. (ii) Of $4{,}721$ ``not what I do'' lead-declines, $94\%$ came from providers whose recorded preferences already excluded that work. While early, we interpret this as an enforcement gap, not a vocabulary gap. Additionally, a review from our category-management team recommended seven point changes to the deployed taxonomy. (iii) About $35\%$ of newly recorded provider limitations arrive through the LLM refinement flow built on the generated catalog. Where no preference was adopted, around 75\% trace to user-experience gaps rather than rejection of a suggested tag ($\sim$$21\%$ explicit rejections). (iv) Finally, churn and refunds seem to cluster amongst pros who had narrow legacy categories that mapped into broad new-catalog occupations, and providers with no legacy-category crosswalk link received $\sim$$35\%$ fewer leads, consistent with the canonical-assignment and parity lessons of \S\ref{sec:operational}. An outcome-level comparison against legacy forms remains future work.

\subsection{Operational Lessons from the Loop}
\label{sec:operational}

\paragraph{Canonical-tag assignment is structurally different from holistic generation.} An earlier version of the pipeline assigned canonical category tags inside the generation prompt. This was migrated to a dedicated mapping-phase LLM call that runs before each mapping batch and injects canonical context. The architectural separation produced both better canonical coverage and more consistent generation (the generator now focuses on holistic tag creation; canonical assignment is a structured decision with its own prompt and its own good/bad examples).
\paragraph{Compound names are not unions.} The canonical assignment rule originally rejected any category whose name contained ``and'' or ``or'' (treating it as a union of distinct services). This collapsed canonical coverage on \emph{Electrical} from 10/10 to 1/10 because categories like \emph{Circuit Breaker Panel or Fuse Box} are compound names for a single service, not unions. The refined rule requires \emph{actual RF Q\&A disambiguation evidence} before rejecting a union category. \paragraph{Stopping is budget-driven.} The current loop stops at a fixed iteration budget (default 5) rather than at a convergence-based criterion. A confidence-interval criterion is dominated in cost by the multi-trial evaluations it would require; across production runs the kept-iteration count typically saturates well before the budget is exhausted (\S\ref{sec:convergence}). The per-iteration early-skip described in \S\ref{sec:loop} step~6 captures most of the available cost saving without committing to a plateau heuristic the implementation does not yet support.

\subsection{Convergence Behavior Across Production Occupations}
\label{sec:convergence}

To characterize how often the loop converges within budget, we aggregate the best-iteration summary row for each of the $132$ production occupations from the autoresearch tracking logs. Table~\ref{tab:convergence} summarizes the distribution.

\begin{table}[t]
\caption{Convergence behavior across all $132$ production occupations. The iteration index reported is the iteration that produced the best kept tag set per occupation (index $0$ denotes occupations whose baseline was never beaten); ``adjusted score'' is $\mathcal{E}\text{ composite} - \text{critic penalty}$. Quartiles use the standard linear-interpolation convention (NumPy \texttt{percentile} default).}
\label{tab:convergence}
\small
\begin{tabular}{lr}
\toprule
Metric & Value (min / q1 / median / q3 / max) \\
\midrule
Iteration index of best tag set & 0 / 1 / 2 / 3 / 3 \\
$\mathcal{E}$ composite of best tag set (/ 15) & 7.69 / 13.32 / 13.84 / 14.20 / 14.90 \\
Critic penalty on best tag set & 0.50 / 2.49 / 3.62 / 4.85 / 9.70 \\
Adjusted score on best tag set (/ 15) & 2.34 / 8.63 / 10.17 / 11.41 / 13.96 \\
Tag count on best tag set & 2 / 17 / 26 / 44.25 / 136 \\
\bottomrule
\end{tabular}
\end{table}

Three observations hold across the full $132$-occupation cohort. (i) The best kept tag set is reached by iteration $3$ for every occupation---$36$ at iteration $1$, $53$ at iteration $2$, $38$ at iteration $3$, and $5$ that never beat their baseline (index $0$)---with none reaching its best after iteration $3$, so the kept-iteration count saturates well before the $5$-iteration budget. (ii) The $\mathcal{E}$ composite is tightly clustered (median $13.84/15$, mean $13.53$, interquartile range $13.32$--$14.20$), but the \emph{adjusted} score spreads much wider (median $10.17$, IQR $8.63$--$11.41$): $19$ of $132$ occupations fall in the $<8$ band that the production verdict thresholds (\S\ref{sec:catalog-quality}) route to review or flag, almost entirely on critic penalty rather than on a weak judge score. (iii) The critic penalty (range $0.50$--$9.70$, median $3.62$) is therefore the dominant source of variance in the adjusted score, consistent with the asymmetric panel design (supplementary material) in which the Adversarial critic alone can contribute up to $2.0$ on the \texttt{critical} risk tier.

\begin{comment}
\hojat{Kartik: added the next paragraph in reponse to loop vs. single-shot question} \shishir{lgtm, thank you!}
\end{comment}
\paragraph{Loop vs.\ single-shot generation.}
Iteration~0 of every production run \emph{is} single-shot generation: the
shared seed prompt $p^{(0)}$ applied once, with no critics and no editor, so
the tracking logs contain the single-pass baseline directly. Across the
132-occupation cohort, the loop lifts the per-occupation $E$ composite from a
median of 11.36 at iteration~0 to 13.84 on the best kept set, a median gain of
$+2.24$ points (IQR 1.60--2.64). It improves 119 of 132 occupations (100 by at
least one point, 87 by at least two); 5 never beat their baseline, and 8 keep
a set with equal-or-lower composite that wins on a smaller critic penalty,
since keeps are decided on the adjusted score. This comparison runs under the
loop's own judge, so it isolates the value of iterate-and-keep over
single-pass generation; whether the judge itself tracks human quality is
assessed separately below.

\paragraph{$\mathcal{E}$-only baseline (vanilla LLM-as-judge contrast).} To isolate the seven-critic panel's marginal contribution at the verdict-decision boundary, we re-bin all $132$ best-iteration rows under an $\mathcal{E}$-only configuration that drops the critic penalty entirely ($\texttt{critic\_penalty} := 0$) and applies the production thresholds (\texttt{auto\_approve} $\geq 11$, \texttt{human\_review} $\geq 8$, \texttt{flag} $< 8$). The two configurations disagree on $84$ of $132$ occupations ($63.6\%$), and every disagreement is in the same direction---the panel's verdict is the same or stricter, the expected sign because the critic penalty is non-negative. The distributions tell the story: $\mathcal{E}$ alone would \texttt{auto\_approve} $125$ of $132$ occupations ($94.7\%$), leaving a human-review queue of only $7$; the panel routes just $47$ to \texttt{auto\_approve}, $66$ to \texttt{human\_review}, and $19$ to \texttt{flag}. The panel's dominant operating mode is therefore to demote tag sets the judge alone would wave through---it is what creates the $85$-occupation human-in-the-loop queue the production system relies on. This is a verdict-bucket comparison on the converged best-iteration tag sets, not a re-run of the loop with the panel disabled; the latter (the panel's effect on the kept-iteration \emph{trajectory}) is the ablation scoped in \S\ref{sec:operational}.

\paragraph{$\mathcal{E}\leftrightarrow$PM-consensus agreement.}
Table~\ref{tab:verdict-confusion} cross-tabulates the model's $E$ verdict against the
production PM-consensus action over the 4{,}112 tags that received both, under
the verdict map Keep$\leftrightarrow$auto-approve,
Update$\leftrightarrow$human-review, Delete$\leftrightarrow$flag. Raw
agreement is 77.2\% (3{,}174/4{,}112), but this is almost entirely a base-rate
effect: the model auto-approves 97.3\% of tags and reviewers keep 79.1\%, so
chance agreement is already 77.2\% and Cohen's $\kappa\approx0$. The discrete
$E$ verdict does not, on its own, predict which tags a human will reject. We
report this negative result deliberately: it is the empirical basis for using
the $E$ composite as a ranking signal inside the loop rather than an
acceptance gate, and for keeping human sign-off mandatory before any catalog
deploys. One caveat is circularity on the legibility rubric, which was
recalibrated from the same PM review (\S\ref{sec:e1}).

\begin{table}[h]
\centering
\small
\caption{Confusion matrix between the model's E1 verdict (columns) and the human
reviewer's action (rows), over $4{,}112$ tags that received both. Categories are aligned
as auto\_approve\,$\leftrightarrow$\,\texttt{Keep},
human\_review\,$\leftrightarrow$\,\texttt{Update}, flag\,$\leftrightarrow$\,\texttt{Delete}.}
\label{tab:verdict-confusion}
\begin{tabular}{@{}lrrrr@{}}
\toprule
 & \multicolumn{3}{c}{\textbf{Model E1 verdict}} & \\
\cmidrule(lr){2-4}
\textbf{Reviewer} & auto\_approve & human\_review & flag & \textbf{Total} \\
\midrule
\texttt{Keep}   & $3{,}164$ & $88$ & $1$ & $3{,}253$ \\
\texttt{Update} & $292$     & $10$ & $0$ & $302$ \\
\texttt{Delete} & $547$     & $10$ & $0$ & $557$ \\
\midrule
\textbf{Total}  & $4{,}003$ & $108$ & $1$ & $4{,}112$ \\
\bottomrule
\end{tabular}
\end{table}

\begin{comment}
\hojat{slight re-framing of the previous paragraph:}\shishir{I updated the prior paragraph to use Hojat's version}
\end{comment}

\section{Discussion}
\label{sec:discussion}
Our contribution is not a single algorithm but a system architecture for reconstructing catalogs in marketplaces that carry substantial legacy structured data, where getting the pro-side schema right has large downstream consequences. The composition we defend has three elements: per-occupation independence, the seven-persona critic panel, and a separate parity-mapping stage. While each module has its own ancestry, what we add is their integration, the scale at which the system runs in production (132 occupations, with new ones onboarded on demand), and the operational lessons that scale has surfaced. We see two clear extension paths as future work. First, the search backend is currently a naive LLM proposal. GEPA-style reflective optimization~\cite{agrawal2025gepa} with execution traces is a natural upgrade once the critic-panel output is structured enough to serve as a reflection signal. Second, the production feedback loop is currently human-in-the-loop (PM comments flow back via the live sheet). Routing matchmaker-side production failures (eg, 41\% missing-canonical-tag filter rate) into the next autoresearch iteration would close the loop between catalog generation and matchmaking outcomes.

\paragraph{Trustworthiness considerations.} The cross-family arrangement described in \S\ref{sec:loop} mitigates self-reinforcement between the generator (GPT-5-4 family) and the seven critics, editor, and parity mapper (Claude Sonnet 4.6 family), but two residual trust gaps remain. (i) Within Claude Sonnet 4.6, the editor and the seven critics share a model family; an editor that is systematically biased in a direction the critics also endorse will not be caught by the loop, even though it would be caught by a cross-family critic seat. A natural mitigation is to host one critic on a third model family (e.g., a Gemini- or Llama-class model) and treat its disagreement as an explicit signal to the editor. (ii) Critic disagreement is currently silent: the editor sees the aggregate adjusted score, not the per-critic decomposition, so a high-variance verdict (e.g., the Adversarial critic flags a critical failure that no other critic surfaces) is averaged into the same scalar as a low-variance verdict. Surfacing per-critic penalties and a disagreement summary as explicit editor inputs is the most direct upgrade to the loop's trustworthiness profile we are aware of, and is a planned next iteration.

\section{Conclusion}

We presented a per-occupation autoresearch system for generating marketplace catalogs, in production at a 132-occupation consumer services marketplace. The contributions we defend are compositional: the unit of optimization (preference-tag catalog within an occupation), the per-occupation parallelism, the seven-persona critic panel with weighted penalties, the legacy-Q\&A-to-modern-tag migration via Step-Back abstraction, and the MEMORY\_ONLY typing primitive. Production deployment at scale surfaced an operational catalog-hygiene finding (three deprecated tags driving 486K filter events) that a global hierarchical build would have masked. The algorithmic ancestry (Karpathy, Self-Refine, Step-Back, Constitutional AI, GEPA) is acknowledged; the integration and the deployment results are what we report.

\bibliographystyle{ACM-Reference-Format}
\bibliography{references}

\end{document}